\documentclass[letterpaper, 10 pt, journal, twoside]{IEEEtran}

\IEEEoverridecommandlockouts                              

\usepackage{cite}
\usepackage{mathtools} 
\usepackage{amsmath} 
\usepackage{amssymb}  
\usepackage{siunitx}
\usepackage{textcomp}
\usepackage{xcolor}
\usepackage[hidelinks]{hyperref}
\usepackage{cleveref}
\usepackage{subcaption}
\usepackage{graphicx}
\usepackage{balance}
\usepackage{booktabs}
\usepackage{framed}
\usepackage{soul}
\usepackage{xcolor}
\usepackage{ulem}
\usepackage{dblfloatfix} 

\DeclareMathOperator*{\argmin}{arg\,min}

\usepackage[colorinlistoftodos,prependcaption,textsize=small]{todonotes}

\usepackage{color}

\newcommand{\rev}[1]{\textcolor{black}{#1}}
\newcommand{\fin}[1]{\textcolor{black}{#1}}

\usepackage[font=small,skip=3pt]{caption}

\IEEEaftertitletext{\vspace{-2em}}

\begin{document}

\title{Sim-to-Real of Soft Robots \\with Learned Residual Physics}

\author{Junpeng Gao$^*$, Mike Y. Michelis$^*$, Andrew Spielberg, and Robert K. Katzschmann%
\thanks{\fin{Manuscript received: February 1, 2024; Revised May 10, 2021; Accepted July 24, 2024. This paper was recommended for publication by Editor Cecilia Laschi upon evaluation of the Associate Editor and Reviewers' comments. Andrew Spielberg is supported by an Amazon Research Award. (\textit{Corresponding author: Robert K. Katzschmann})}
} 
\thanks{Junpeng Gao is with the Soft Robotics Lab, Department of Mechanical and Process Engineering, ETH Zurich, Switzerland
        {\tt\footnotesize \href{mailto:jungao@ethz.ch}{jungao@ethz.ch}}}%
\thanks{Mike Y. Michelis and Robert K. Katzschmann are with the Soft Robotics Lab and ETH AI Center, ETH Zurich, Switzerland
        {\tt\footnotesize \{\href{mailto:michelism@ethz.ch}{michelism}, \href{mailto:rkk@ethz.ch}{rkk}\}@ethz.ch}}%
\thanks{Andrew Spielberg is with Harvard University, USA
        {\tt\footnotesize \href{mailto:aespielberg@seas.harvard.edu}{aespielberg@seas.harvard.edu}}}%
\thanks{* Equal contribution.}%
\thanks{\fin{Videos, code, and data are available on our website at: \texttt{\href{https://srl-ethz.github.io/website_residual_physics/}{https://srl-ethz.github.io/website\_residual\_physics/}}}}%
\thanks{Digital Object Identifier (DOI): see top of this page.}
}

\markboth{IEEE Robotics and Automation Letters. Preprint Version. Accepted July, 2024}{Gao \MakeLowercase{\textit{et al.}}: Sim-to-Real of Soft Robots with Learned Residual Physics} 


\maketitle


\begin{abstract}

Accurately modeling soft robots in simulation is computationally expensive and commonly falls short of representing the real world. This well-known discrepancy, known as the sim-to-real gap, can have several causes, such as coarsely approximated geometry and material models, manufacturing defects, viscoelasticity and plasticity, and hysteresis effects. Residual physics networks learn from real-world data to augment a discrepant model and bring it closer to reality. Here, we present a residual physics method for modeling soft robots with large degrees of freedom. We train neural networks to learn a residual term --- the modeling error between simulated and physical systems. Concretely, the residual term is a force applied on the whole simulated mesh, while real position data is collected with only sparse motion markers. The physical prior of the analytical simulation provides a starting point for the residual network, and the combined model is more informed than if physics were learned \textit{tabula rasa}. We demonstrate our method on \textit{1)} a silicone elastomeric beam and \textit{2)} a soft pneumatic arm with hard-to-model, anisotropic fiber reinforcements.  Our method outperforms traditional system identification up to \SI{60}{\%}. We show that residual physics need not be limited to low degrees of freedom but can effectively bridge the sim-to-real gap for high dimensional systems. 

\end{abstract}



\section{Introduction}
\IEEEPARstart{W}{e}  present a data-driven approach for reducing the sim-to-real gap in soft robotics.  Despite soft robots' promise in solving tasks that are difficult for rigid robots to solve (\textit{e.g.}, delicate manipulation \cite{junge2023lab2field} and biomimicry  \cite{katzschmann2018exploration}), modeling soft robots remains computationally expensive and physically inaccurate. This challenge hinders the application of computational methods for downstream tasks such as optimal control and design. By providing a generic means to improve simulation accuracy that is system agnostic, we can unlock applications across the diverse zoo of soft robotics.

Simulators that model soft robots commonly have limited options for fitting physical parameters. Once a material model is chosen, only a small set of parameters, such as material density, stiffness, compressibility, friction, and damping, can be tweaked to adjust the behavior of a soft body. In most practical applications, such parameter tweaking is sufficient to better match the simulated model with its real-world counterpart; this process is referred to as system identification (SysID) \cite{zhang2022sim, dubied2022sim, ljung1998system}. However, suppose the simulator's physics does not match the real world for reasons other than parameter mismatch, for example, incorrect material model, overly coarse discretization in time or space, or simulation artifacts (such as locking in finite element methods). In this case, the expressivity of the simulator may not suffice to cover the real-world dynamics.  

\begin{figure}[!tb]
    \centering
    \includegraphics[width=0.95\columnwidth]{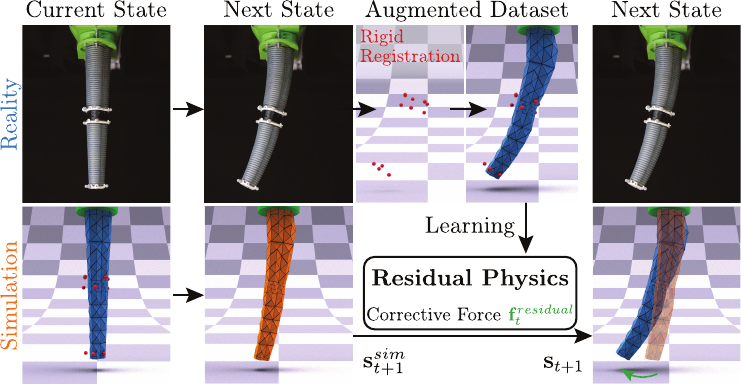}
    \caption{Overview of the residual physics pipeline for high dimensional systems, demonstrated with a soft robotic arm. The learned residual force compensates for state-to-state prediction errors, such that sparse motion markers in simulation match those in reality.}
    \label{fig:overview}
    \vspace{-2em}
\end{figure}

We propose to combine deformable body simulators with data-driven auxiliary models, as a means of reducing the sim-to-real gap. We see this approach as a viable alternative to tediously modeling every possible aspect of every continuum-bodied robot, which would only grow with the creation of further soft robotic systems.
Our \textit{\rev{sparse} residual physics learning framework} is a hybrid formulation taking advantage of a differentiable Finite Element Method (FEM) simulator and deep learning. Our framework learns a residual body force on the soft structure that captures the difference between the simulator and the real-world, \textit{directly} minimizing the sim-to-real error. This formulation combines the qualitative priors of a coarse simulator, \textit{e.g.}, direction of bending and approximate magnitude of deformation, with the fine-tuning derived from real-world data. \rev{Unlike previous residual learning approaches (\textit{e.g.} \cite{ajay2018augmenting, zeng2020tossing}) we operate in a regime of sparse observation data, since continuum structures that cannot be fully sensorized are inherently partially observable. Our approach regularizes on system dynamics to create physically reasonable candidates for target learned motions and provides a means for a simulator to capture unmodeled robot dynamics and unseen settings.}

We provide a full pipeline\footnote{All code and data used in this paper are available at \url{https://github.com/srl-ethz/residual_physics_sim2real}} from spatially sparse data to dense residual force estimation (visualized in \Cref{fig:overview}) and apply our approach to both software and hardware experiments. Our approach is easy to apply in practice and can improve soft robotics engineering workflows through more reliable modeling. In summary, we contribute:

\begin{enumerate}
    \item \emph{Residual physics learning framework for soft robotics.} We design a hybrid learning framework that speeds up simulation while increasing simulation accuracy in both sim-to-sim and sim-to-real settings.
    
    \item \emph{Dense residuals from sparse observations.} We propose a data pre-processing method to build an augmented dataset from sparse real-world data (markers \mbox{$\sim10^1$}) for learning residual physics on the discretized geometry of deformable bodies (degrees of freedom \mbox{$\sim10^3$}). 
    
    \item \emph{Overcoming shortcomings of system identification.} We benchmark our framework on dynamical high-dimensional systems, such as passive and actuated soft robots, and show that even optimal tuning of physical simulation parameters falls short in accuracy compared to our approach.
\end{enumerate}

\section{Related Work}

Much effort has gone into developing more efficient and accurate simulators for applications of deformable systems over the years \cite{muller2007position, bouaziz2014projective, macklin_constraint-based_2021, Schneider:2019:PFM, Li2020IPC, gazzola2018forward, skouras2013computational, 8793704, cangan2022model, katzschmann2019dynamically}. The consistent trend, however, has been a trade-off between speed and accuracy. For the computer graphics community, a visually plausible simulation for soft bodies usually suffices for applications in animation and digital gaming.  Fidelity is sacrificed for speed, enabling real-time simulation of deformable virtual characters \cite{muller2007position, bouaziz2014projective, macklin_constraint-based_2021}. Within the soft robotics community, much more emphasis is placed on the physical accuracy of the model to match real-world experiments \cite{gazzola2018forward, skouras2013computational, 8793704, cangan2022model, katzschmann2019dynamically}.  Here, simulation is particularly important to test design performance without the labor of physical manufacturing and to derive optimal controllers on real-world systems. 

Matching real-world experiments, or in other words, closing the sim-to-real gap, is traditionally done through SysID, where a set of simulation parameters is tuned.  In soft robotics, these parameters often include material characteristics such as stiffness and density. Gradient-free approaches are often sample-inefficient and hence costly to run for larger sets of parameters \cite{du2021_diffpd, zhang2022sim, hu2019chainqueen}. For this type of inverse problem, gradient-based optimization through differentiable simulation frameworks have offered significant improvement in convergence time \cite{du2021_diffpd, zhang2022sim, du2021underwater, dubied2022sim, jatavallabhula_gradsim_2021, ma2022risp, hu2019chainqueen}. Differentiable simulations provide analytical gradients for any measurable quantity of a simulation with respect to any simulation parameter; such gradients are useful for inverse problems such as trajectory-matching and system identification. While most methods estimate the state of the system through sparse motion markers, recent methods have integrated differentiable rendering pipelines to allow direct parameter matching from video data \cite{jatavallabhula_gradsim_2021, ma2022risp}.

Since one notable limitation of SysID is the need to re-run the procedure when characteristics about the system change, one approach is to learn a generalizable mapping from one environment to another, tuning these SysID parameters iteratively in a fast and sample-efficient manner \cite{allevato2020tunenet}. Yet an overarching challenge for bridging the sim-to-real gap remains the many real-world physical phenomena that are not explicitly modeled by the underlying simulations, ranging from electric actuator dynamics \cite{hwangbo2019learning} to flying robot aerodynamics \cite{kaufmann2023champion}. This need for a more generalizable model for matching simulation and reality has given rise to the field of residual learning \fin{\cite{ajay2018augmenting, zeng2020tossing, heiden2021neuralsim, golemo2018sim, hwangbo2019learning, kaufmann2023champion}}. Instead of learning the full system dynamics, residual physics methods use deep neural networks to learn only an error correction between an analytical simulator and real-world physics. Previous work, however, has only worked with low-dimensional state spaces, and it has yet to be scaled up to the high degree-of-freedom meshes used in soft-body simulations \fin{\cite{chin2020machine}}. 

A valid alternative to the previous hybrid residual physics simulations would be fully end-to-end learning-based simulators \cite{battaglia2016interaction, li2018learning, pfaff2020learning, laschi2023learning}. Although computationally efficient at inference time, these data-driven methods lack generalizability and robustness \cite{ajay2018augmenting, heiden2021neuralsim}. This fragility can be ameliorated by including constrained neural networks into hybrid simulations \cite{ma2023learning}; the analytical solvers within such hybrid methods guide the solution along a physically plausible prediction. However, such an approach is confined to a PDE's structure and cannot handle unmodeled phenomena.

\section{Simulation Preliminaries}
\label{preliminaries}
We simulate our soft robots using DiffPD~\cite{du2021_diffpd}, a differentiable FEM simulator based on projective dynamics. Each robot is discretized as having $N$ nodes, where we denote the position and velocity of nodes at time step $t$ with $\mathbf{q}_t \in \mathbb{R}^{N\times3}$ and $\mathbf{v}_t\in \mathbb{R}^{N\times3}$ respectively. Using implicit Euler, the simulation is integrated in time according to Newton's second law of motion over fixed time interval $h$. The resulting equations to the discretized dynamical system can be formulated as
\begin{equation}\label{eq1}
\begin{aligned}
& \mathbf{q}_{t+1}=\mathbf{q}_t + h \mathbf{v}_{t+1} \\
& \mathbf{v}_{t+1}=\mathbf{v}_t + h \mathbf{M}^{-1}\left[\mathbf{f}^{\mathrm{int}}\left(\mathbf{q}_{t+1}\right)+\mathbf{f}_t^{\mathrm{ext}}\right]
\end{aligned}
\end{equation}
where $\mathbf{M}$ is the mass-matrix, $\mathbf{f}^{\mathrm{int}}$ accounts for the sum of internal forces and $\mathbf{f}_t^{\mathrm{ext}}$ for the sum of external forces. This results in a system of equations solved as a sometimes numerically and physically stiff optimization problem. For the sake of clarity, we simplify this forward solve as $\mathbf{q}_{t+1}, \mathbf{v}_{t+1} = \texttt{Sim} \left( \mathbf{q}_{t}, \mathbf{v}_{t}, \mathbf{f}_t^{\mathrm{ext}} \right)$.

\begin{figure*}[!tb]
\centering
  \includegraphics[width=\textwidth]{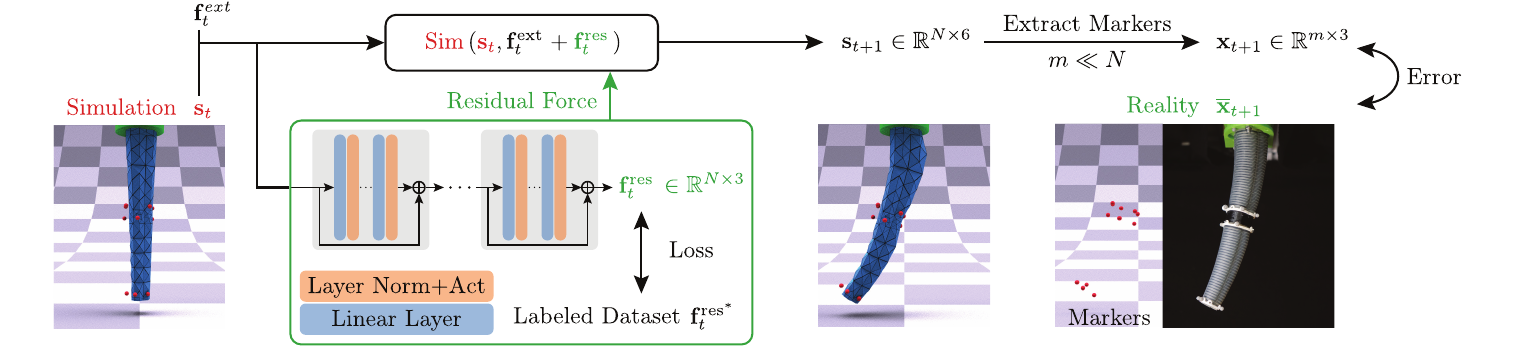}
  \caption{Pipeline of how the residual physics forces $\mathbf{f}_t^{\mathrm{res}}$ compensate the erroneous simulated next state $\mathbf{s}_{t+1}$ to match the real observed marker state $\overline{\mathbf{x}}_{t+1}$. Our state $\mathbf{s}_{t}$ is defined by position $\mathbf{q}_{t}$ and velocity $\mathbf{v}_{t}$, from which we extract the motion markers $\mathbf{x}_{t}$ on the simulated mesh. The residual forces are predicted by a neural network given state and external force $\mathbf{f}_t^{\mathrm{ext}}$  information (such as pressure actuation) as input. This network is trained on a labeled augmented dataset of residual forces $\mathbf{f}_t^{\mathrm{res}^*}$, collected through gradient-based optimization in our differentiable simulation.}
  \label{fig:flowchart}
  \vspace{-1em}
\end{figure*}

The deformable structures used in this paper, a passive beam and a pneumatic arm, are both made from highly deformable silicone elastomers, namely Smooth-On Dragon Skin 10 with Shore Hardness 10A. As is common \cite{du2021_diffpd}, we make a few modeling simplifications to aid elastic simulation stability. First, we use a corotational linear elastic material model. Second, silicone is typically assumed to be completely or nearly incompressible \cite{dubied2022sim}, \textit{i.e.}, Poisson's ratio $\nu = 0.5$ or $0.499$. However, such Poisson's ratios are both unstable and time-stepping can take longer to converge when simulated; we set $\nu = 0.45$. We expect our residual physics (ResPhys) framework to be able to compensate for both the material model and compressibility assumptions. We roughly use manufacturer-provided values for material density and Young's modulus ($\rho =$ \SI{1070}{\kilo\gram\per\cubic\meter} and $E = \SI{215}{\kilo\pascal} $), which have less impact on time-stepping convergence \cite{ljung1998system}.

\section{Method}
The objective of our framework is to learn a mapping from the state $\mathbf{s}_t$, consisting of positions $\mathbf{q}_t$ and velocity $\mathbf{v}_t$, and the action of the soft robot at time step $t$ to an external residual body force $\mathbf{f}^{\mathrm{res}}_t$. 
The design of our framework is based on the assumptions that \textit{1}) an external residual force in \Cref{eq1} can compensate for the residual dynamics of the simulated deformable objects, and \textit{2}) if the simulator can predict the next state precisely, the residual force at each time step will follow similar distributions. This assumption will make the chosen neural network smaller and more efficient. 

In this section, we describe our procedure for leveraging data to overcome the sim-to-real gap in soft robot modeling, and illustrate the pipeline in \Cref{fig:flowchart}. We begin by considering a simplified \textit{sim-to-sim} setting, in which full state knowledge is available; we then relax this knowledge requirement to arrive at our method for \textit{sim-to-real}, in which only partial information with measurement error is given about the physical world. Next, we describe the system identification we use as a baseline. Lastly, we describe the neural network used for the residual physics learning, how it is trained, and how we quantify the prediction performance.

\subsection{Sim-to-Sim Setting}
\label{method:sim2sim}

By removing the influence of potential fabrication and measurement errors, a \textit{sim-to-sim} setting enables us to rigorously validate and refine our framework at a relatively low cost. Such a setting can provide essential insights and help pave the way for real-world experimentation.

We define a function $\texttt{Sim}_n$ that takes as input state $\mathbf{s}_t$ and outputs position coordinates $\mathbf{q}_{t+1}$; the subscript $n$ denotes a particular parameterization of DiffPD. We define the state $\mathbf{s}_t$ as a concatenation of $\mathbf{q}_t$ and $\mathbf{v}_t$ at time step $t$.  We parameterize two differentiable simulators $\texttt{Sim}_1, \texttt{Sim}_2$ with different parameter configurations for sim-to-sim experiments. We aim to match the dynamics of $\texttt{Sim}_2$ by injecting external residual forces generated from our framework into $\texttt{Sim}_1$. Sim-to-sim scenarios provide a privileged, fully-known robot state at a low cost, as we can easily obtain the positions and velocity at each degree of freedom (DoF). We collect a series of ground truth motion data by running $\texttt{Sim}_2$ offline. 

We decompose the residual learning problem into two separate steps to ease the computational cost of rerunning each specific step. First, we leverage the differentiable property of the simulator to optimize for a dataset of external forces in $\texttt{Sim}_1$ that makes the simulated motion trajectory best match the results of $\texttt{Sim}_2$. Second, we perform supervised learning on this optimized dataset of external forces to create a neural network mapping between soft body state and residual forces. The network takes an input of the state $\mathbf{s}_t$ and, if the structure is actuated, the actuation forces, which are represented as external forces $\mathbf{f}^{\mathrm{ext}}_t$ (such as pneumatic pressure forces). The output is the residual force $\mathbf{f}^{\mathrm{res}}_t$ that helps the simulator correct the prediction. Note that gravity is applied separately \rev{in all simulations, but it is not considered as part of the external force input for the network.}

The first step can be formalized as follows:
\begin{equation}
    \mathbf{f}^{\mathrm{res}^*}_t = \argmin\limits_{\mathbf{f}_t} \left\| \texttt{Sim}_1(\mathbf{s}_{t}, \mathbf{f}^{\mathrm{ext}}_t+\mathbf{f}_t) - \overline{\mathbf{q}}_{t+1} \right\|_2^2 + \lambda \left\| \mathbf{f}_t \right\|_2^2
\label{eq:sim2sim_op}
\end{equation}
where $\mathbf{f}_t$ is the residual force we aim to optimize at time step $t$, and $\overline{\mathbf{q}}_{t+1}$ is the ground truth full state from $\texttt{Sim}_2$ at the next time step. We incorporate $L_{2}$ regularization with a weight $\lambda$ \rev{so that the residual forces can help predict an accurate next state while having a small magnitude, aiding simulator stability.}
We leverage the differentiability from the simulator to solve the above optimization problem using an efficient \mbox{L-BFGS-B} minimization. In the first time step, we initialize a random residual force $\mathbf{f}_1 \sim \mathcal{N}\left(0, 10^{-4} \right)$. In subsequent time steps, we use the preceding $\mathbf{f}_{t}$ as an initial guess for the ongoing step to solve the problem iteratively. 

After we build a dataset of various trajectories over a fixed number of time steps each, we perform mini-batch training with batch-size $M$ of the neural network $\mathcal{NN}$ with weights $\Theta$ based on the loss function:
\begin{gather}
\label{eq:sim2sim_loss}
\begin{aligned}
    &\mathcal{L} := \sum_{i \in \mathrm{batch}} \left\| \mathcal{NN}(\mathbf{s}_i, \mathbf{f}^{\mathrm{ext}}_i; \Theta) - \mathbf{f}^{\mathrm{res}^*}_i \right\|_2^2 + \lambda \left\| \Theta \right\|_2^2
\end{aligned}
\end{gather}
When we pass the state into the neural network, as a pre-processing step, we subtract the undeformed static state from the position $\mathbf{q}$. After each epoch, we run validation with the same loss function \Cref{eq:sim2sim_loss} on the validation set and save the model with the smallest validation error. This two-step training method can be thought of as a student-teacher formulation, in which residual forces are generated from privileged information, from which a residual network then learns with no privileged knowledge.

We test our trained model on $R$ trajectories and evaluate the performance on $T$ timesteps for each trajectory with:
\begin{equation}
\label{eq:sim2sim error}
\mathcal{E}_q := \frac{1}{R \cdot T \cdot N}\sum_{j=1}^R\sum_{t=1}^T\sum_{i=1}^N \left\|\mathbf{q}^j_{t,i}-\overline{\mathbf{q}}^j_{t,i}\right\|_{2}
\end{equation} 
where $\mathbf{q}_{t,i}^j$ and $\overline{\mathbf{q}}_{t,i}^j$ are the simulated position coordinates at vertex $i$ of $\texttt{Sim}_1$ and $\texttt{Sim}_2$ respectively, at the $j$-th trajectory and time step $t$.

\subsection{Sim-to-Real Setting}
\label{method:sim2real}

Under a sim-to-real scenario, measuring full-state information is impractical, as it would require sensorizing every point of the target object throughout its motion.  In the sim-to-real scenario, we record the motion of our robot with a marker-based motion capture system, which provides us with sparse partial information, but this data may lie in a different coordinate frame than our simulation environment. Therefore, we perform rigid registration to transform the frame of reference of the raw measurement data. We first use the undeformed state marker measurements of the object to define simulated marker locations. Subsequently, we estimate an optimal rotation matrix and translation vector between the measured markers and the recorded data at the corresponding undeformed state. Finally, we apply this transformation to the collected data and interpolate the transformed markers with 
\begin{equation}
\label{eq:marker_interpolation}
    \mathbf{x}(\mathbf{q}) = \mathbf{\alpha}^T\mathbf{e}(\mathbf{q}) + s\mathbf{n}
\end{equation}
where $\mathbf{n}$ is the unit normal vector to the closest surface mesh element, $s$ is the distance between the transformed marker to the element, $\mathbf{\alpha}$ is a barycentric coordinate vector w.r.t. the neighboring surface element nodes $\mathbf{e}$. During forward passes of the simulation we can compute simulated marker positions by interpolating the corresponding surface mesh element with \Cref{eq:marker_interpolation}.

Next, we create our augmented dataset of reconstructed full-state information from the sparse partial information, similarly to \Cref{eq:sim2sim_op}:
\begin{equation}
    \mathbf{f}^{\mathrm{res}^*}_t = \argmin\limits_{\mathbf{f}_t} \left\| \mathbf{x}\left( \texttt{Sim}(\mathbf{s}_{t}, \mathbf{f}^{\mathrm{ext}}_t+\mathbf{f}_t) \right) - \overline{\mathbf{x}}_{t+1} \right\|_2^2 + \lambda \left\| \mathbf{f}_t \right\|_2^2
\label{eq:sim2real_op}
\end{equation}
where $\mathbf{x}$ returns the simulated markers from the returned full state of the simulator, and $\overline{\mathbf{x}}_{t+1}$ is the transformed real markers at time step $t+1$. 

With our residual force dataset now well-defined in the real-world scenario, we have reduced our sim-to-real problem to the same setting as that of the sim-to-sim problem.  As such, we solve the simplified problem directly, and train a residual model with the loss as described in \Cref{eq:sim2sim_loss}. As before, we assess the model's performance on the validation set after each training epoch and save the model with smallest validation error.
We evaluate the saved model across $R$ test trajectories and compute the mean rollout error between simulated markers and their real counterparts. We rollout the hybrid simulation with our trained network in an auto-regressive manner; starting from the same initial state as the ground truth motion, we feed in the previous position and velocity solutions from the simulation, and add the predicted residual physics to the next forward pass of the simulation. The error is computed based on $m$ markers as follows:
\begin{equation}
    \mathcal{E}_x := \frac{1}{R \cdot T \cdot m}\sum_{j=1}^R\sum_{t=1}^T\sum_{i=1}^m \left\| \mathbf{x}^j_{t,i}-\overline{\mathbf{x}}^j_{t,i} \right\|_2
\label{eq:marker errors}
\end{equation}
where $\mathbf{x}^j_{t,i}$ denotes the $i$-th marker position at the $j$-th trajectory and time step $t$.

\subsection{Residual Physics Learning}

We divide the network architecture into blocks that each contain Multilayer Perceptron (MLP), with skip connections added between blocks. A layer-normalization and Exponential Linear Unit (ELU) activation is applied to the output of each linear layer except the last, and the network is trained based on \Cref{eq:sim2sim_loss}. The input layer takes in the positions, velocities, and actuation forces of the simulated system; note that for unactuated systems, the length of $\textbf{f}^{\mathrm{ext}}$ will be zero. An overview of the residual physics network is shown in \Cref{fig:flowchart}. We standardize each spatial dimension $(x, y, z)$ of our datasets for positions $\mathbf{q}$, velocities $\mathbf{v}$, and actuation forces $\mathbf{f}^{\mathrm{ext}}$ to zero mean and unit standard deviation based on the training data, then apply the same standardization during validation and testing.


We run the sim-to-sim network on the datasets described in \Cref{table:sim2sim_training_params} and the sim-to-real network based on the datasets described in \Cref{table:sim2real_params} for 1000 epochs. We set the input and output size of each block the same as our network output size. We optimize the hyperparameters of the network through a Bayesian optimization over the batch size, learning rate, scheduler gamma, number of hidden sizes, forward layers per block, and the number of blocks. The validation loss is used to determine which network architecture to keep during this hyperparameter optimization. \rev{Details on hyperparameter ranges can be found in the code repository.}


\begin{figure}[!t]
    \centering
    \vspace{0.1em}
    \includegraphics[width=\linewidth]{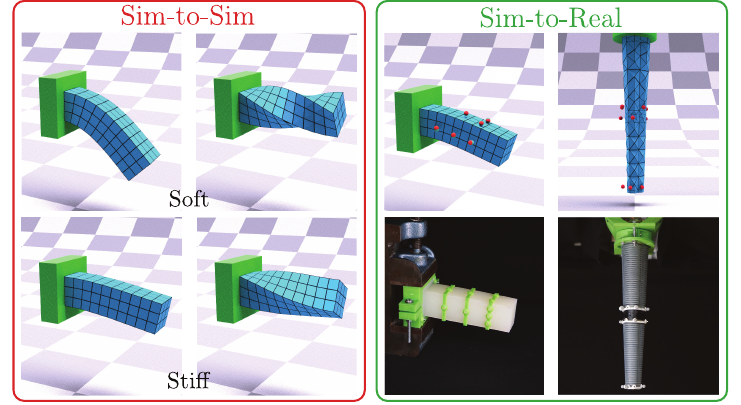}
  \caption{Sim-to-sim experiments include oscillating and twisting beams, where we either apply a weight at the tip or twist the beam and release this constraint to observe a desired motion trajectory. The sim-to-real experiments show the same passive oscillating beam and a pneumatic soft arm as an actuated robot.}
  \label{fig:initials}
\end{figure}

\subsection{System Identification as Baseline}

In the sim-to-real setting, we run SysID as a baseline to our ResPhys approach. We optimize the Young's modulus $E$ and Poisson's ratio $\nu$ to minimize the following objective function on the training dataset with $R$ trajectories, based on the distance between transformed real markers $\overline{\mathbf{x}}_t$ and simulated markers $\mathbf{x}_t$:
\begin{equation}
    \mathcal{L}_s := \frac{1}{2R}\sum_{j=1}^R \sum_{t=1}^T \left\| \mathbf{x} \left( \texttt{Sim}(\mathbf{s}_{t}^j,\mathbf{f}^{\mathrm{ext},j}_t; E, \nu) \right) -\overline{\mathbf{x}}^j_t \right\|^2_2
\label{eq:sysid}
\end{equation}

\subsection{Sim-free Method as Baseline}
\rev{
We use the network architectures described before in residual physics, but instead of predicting corrective forces, this data-driven baseline predicts the next state directly; a mapping from $\mathbf{s}_t$ to $\mathbf{s}_{t+1}$. This method serves as a direct baseline to showcase the effectiveness of our model-based residual physics. We similarly optimize the sim-free hyperparameters, and test the networks with lowest validation error.
}

\section{Results}

We discuss two continuum deformable structures in the sim-to-real setting: A clamped soft beam and a soft robotic arm called SoPrA\cite{Toshimitsu_2021}. The beam has measurements $\SI{10}{cm} \times \SI{3}{cm} \times \SI{3}{cm}$, and the arm is $\SI{30}{cm}$ in length, with an outer tip diameter of $\SI{3}{cm}$. SoPrA is made with the same silicone elastomer as the soft beam, and it has six fiber-reinforced chambers. Fiber reinforcement structures are useful in preventing excessive inflation of fluidic actuators but introduce composite material anisotropy, making it an interesting application domain for our residual physics framework.

All results were performed on a computer with a 32-Core AMD Ryzen Threadripper 3970X CPU and an RTX 3090 GPU. The GPU was used for training, while the CPU was used for all inference scenarios to minimize overhead since DiffPD is a pure CPU-based simulation framework.

\subsection{Sim-to-Sim for Soft Beam}
\label{sec:sim2simbeam}
In our first experiment, we test to see if two soft clamped beams, one with ``correct'' and one with ``incorrect'' material parameters as shown in the \Cref{table:sim2sim_prams}, can be translated between each other \textit{via} residual physics. Since we employ a linear corotational model, the material is not particularly complex, and the problem should be tractable.  However, this setting also allows us to analyze our algorithm with granularity.

\begin{table}[h] 
  \centering
  \vspace{0.5em}
  \setlength{\tabcolsep}{4.5 pt}
  \caption{Parameter configurations for sim-to-sim beams.}
  \scalebox{1.}{
  \begin{tabular}{lcc}
  \toprule
  Parameters & Incorrect beam  & Correct beam \cite{dubied2022sim}\\ 
  \midrule
  DoFs  & 528 & 528\\
  Poisson ratio & 0.45 &0.499\\
  Young's modulus  & \SI{215}{\kilo\pascal} & \SI{264}{\kilo\pascal}\\ 
  \bottomrule
  \label{table:sim2sim_prams}
  \end{tabular}
  }
  \vspace{-2em}
\end{table}

We consider two motion patterns for the simulated clamped beam displayed in \Cref{fig:initials}. In the first one, we apply a force to the tip of the beam, wait for it to reach a steady state, and release it to observe the oscillations. In the second pattern, we twist the beam at varying angles within the range of $\left[\frac{\pi}{6}, \pi\right]$, then release the beam. We apply the same tip forces for the oscillating beam as the weights we use in the real experiments in \Cref{sec:sim2real beam}. \rev{We choose $\lambda =10^{-4}$ in \Cref{eq:sim2sim_op} and \Cref{eq:sim2real_op} when we optimize residual forces for all the beam experiments.} We initialize our neural network following the discussion of \Cref{preliminaries}.


In \Cref{table:sim2sim_training_params}, we report the mean rollout error in \Cref{eq:sim2sim error} for those test trajectories. \rev{As expected, both purely data-driven and residual physics approaches can learn these simple dynamics effectively, though residuals are easier to learn and result in more accurate final trajectories.}

\begin{table}[h]
  \centering
  \setlength{\tabcolsep}{2.5 pt}
  \caption{\rev{Sim-to-sim experimental configurations and results.}}
  \scalebox{1.}{
  \begin{tabular}{lcc}
  \toprule
   & Oscillating Beam & Twisting Beam\\ 
  \midrule
  Step size (\SI{}{s}) & 0.01 & 0.01 \\
  Time Steps & 150 & 100\\
  Training Trajectories & 9 & 10\\    
  Validation Trajectories  & 2 & 2\\
  Testing Trajectories & 5 & 8\\
  \midrule
  Simulation Error (\SI{}{mm}) & $4.488 \pm 1.468$ &$4.362 \pm 1.541$ \\
  \rev{SimFree Error (\SI{}{mm})} & $0.035 \pm 0.070$ & $0.038 \pm 0.147$ \\
  ResPhys Error (\SI{}{mm}) & $\textbf{0.004} \pm \textbf{0.010}$ & $\textbf{0.004} \pm \textbf{0.011}$\\
  \bottomrule
  \label{table:sim2sim_training_params}
  \end{tabular}
  }
  \vspace{-2em}
\end{table}

\subsection{Full State Reconstruction from Sparse Markers}
As we collect sparse partial information from a marker-based system, we design an experiment in simulation to investigate how the number of markers influences the reconstruction of the full state information. 
The previous experiment in \Cref{sec:sim2simbeam} was performed using full state information; in the following, we assume only to have access to a subset of surface vertices that artificially represent the motion markers we would use in the real world.

\begin{figure}[!b]
    \centering
    \includegraphics{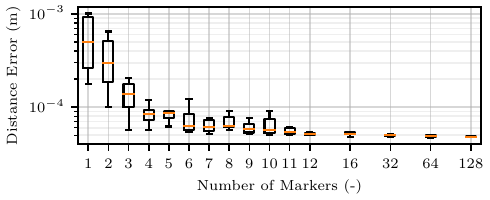}
    \caption{Box plot of displacement error in \Cref{eq:sim2sim error}, varying the number of markers that are available for the residual forces in \Cref{eq:sim2real_op}. The orange line is the median of 10 samples, and the box extends from the lower to the upper quartile of the samples.}
    \label{fig:MarkersNum}
\end{figure}

We choose a single trajectory with a tip force caused by a \SI{50}{g} mass for the oscillating beam. We solve \Cref{eq:sim2real_op} for an increasing number of randomly sampled motion markers, starting from a single marker and ending at 128 markers, where the total number of surface vertices for this mesh is 140. We uniformly randomly sample each subset of markers 10 times and optimize \Cref{eq:sim2real_op} from scratch.

We observe in \Cref{fig:MarkersNum} that the median error drops below \SI{0.1}{mm} starting from 4 markers, though the spread stays high until 10 markers. This sim-to-sim experiment was designed to verify that our pipeline not only works for full-state information, but also when only sparse observations are available, such as in the sim-to-real scenarios that follow. 

\subsection{Sim-to-Real for Soft Beam}
\label{sec:sim2real beam}

We set up a simulator for the beam following the parameters in \Cref{table:sim2real_params}. Our objective is to match the position of simulated markers to the corresponding real markers. To collect motion data of the beam, we attach a set of 17 known weights ranging between \qtyrange{50}{210}{\gram} to the tip of the beam. After the beam reaches a steady state under the weight, we release it and let it oscillate freely. To track the motion of the soft structures, we use a motion capture system (Miqus M3, Qualisys) that runs at \SI{100}{\hertz}. 

\begin{table}[!htb]
  \centering
  \setlength{\tabcolsep}{4.5 pt}
  \caption{Sim-to-real experimental configurations.}
  \scalebox{1.}{
  \begin{tabular}{lcc}
  \toprule
  Model & Beam & SoPrA\\ 
  \midrule
  DoFs  & 528 & 1482\\
  Poisson ratio & 0.45 &0.45\\
  Young's modulus (\SI{}{\kilo\pascal})  & 215 & 215\\ 
  Step size (\SI{}{s}) & 0.01 &0.01 \\
  Markers & 10 & 12\\
  Time Steps & 140 & 1000\\
  Training Trajectories & 9 & 35\\    
  Validation Trajectories  & 2 & 5\\
  Testing Trajectories & 5 & 10\\
  \bottomrule
  \label{table:sim2real_params}
  \end{tabular}
  }
  \vspace{-2em}
\end{table}

\begin{figure}[!t]
    \centering
    \includegraphics{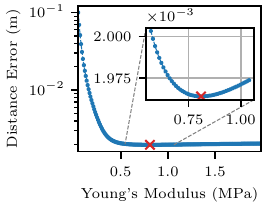}
    \caption{System identification performed with grid search.}
    \label{fig:sim2simbeam}
    \vspace{-0.5em}
\end{figure}

We run SysID with a Youngs' modulus within \qtyrange{0.05}{2}{\MPa} and Poisson's ratio within \qtyrange{-0.999}{0.499}{}. Our final optimized parameters converge to \SI{810}{\kilo\pascal} and 0.499, respectively. The Young's modulus value is much larger than the value reported by the manufacturers. Hence to validate and understand our findings, we run a system identification grid search at resolution \SI{10}{\kilo\pascal} as depicted in \Cref{fig:sim2simbeam} on the left. The optimal Young's modulus value is obtained at \SI{810}{\kilo\pascal}, aligning with our gradient-based optimization result and highlighting the flat nature of the objective landscape. The optimal value is far from ground truth, highlighting the mismatch between the simulation and the physical world.

Building the augmented dataset described in \mbox{\Cref{method:sim2real}} requires us to know the initial state of the beam under weights, so for each weight, we start from an undeformed state $\mathbf{s}_1$ and optimize a series of virtual forces $\mathbf{f}^v_t$ such that after $T_v$ steps we match the initial marker positions $\overline{\mathbf{x}}_{\mathrm{init}}$:
\begin{equation}
    \mathcal{L}_{\mathrm{init}} := \sum_{t=1}^{T_v} \left\| \textbf{x} \left( \texttt{Sim}(\mathbf{s}_{t}, \mathbf{f}^v_t) \right) - \overline{\mathbf{x}}_{\mathrm{init}} \right\|_2^2 + \lambda \left\| \mathbf{f}^v_{1...T_v}  \right\|_2^2
\end{equation}

$T_v$ should be chosen to reach a steady state, which we set to $T_v = 140$. This is the static analog to \Cref{eq:sim2real_op}. The solution to the optimization problem $\mathbf{s}_{T_v}$ is our ground truth's initial position, from which we build the augmented dataset and train the network as described in \Cref{method:sim2real}.

\begin{figure}[!b]
  \begin{subfigure}[t]{0.495\columnwidth}
  \centering
    \includegraphics{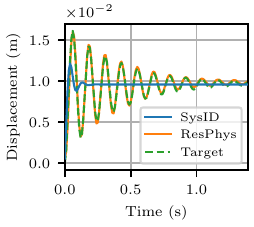}
  \end{subfigure}
  \hfill 
  \begin{subfigure}[t]{0.495\columnwidth}
  \centering
    \includegraphics{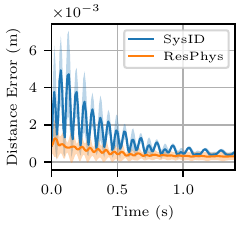}
  \end{subfigure}
  \caption{Sim-to-real results on the soft beam. (Left) Single test trajectory showing displacement in axis of oscillation averaged over motion markers. (Right) Mean and standard deviation of errors on all test trajectories plotted over time.}
  \label{fig:sim2real_beam}
\end{figure}

\begin{table}[h]
  \centering
  \setlength{\tabcolsep}{3.5 pt}
  \caption{\rev{Mean rollout error between simulated and real motion markers for passive and actuated soft structures.}}
  \scalebox{1.}{
  \begin{tabular}{lcc}
  \toprule
  Experiments & Beam Error (\SI{}{mm}) & SoPrA Error (\SI{}{mm})\\ 
  \midrule
  Original & $5.46 \pm  1.78 $ & $7.50 \pm 3.69$ \\
  SysID  & $1.32 \pm  1.38 $ & $7.31  \pm 3.54  $\\
  \rev{SimFree}  & $0.63 \pm  0.49 $ & $8.35  \pm 4.03 $\\
  ResPhys   &  $\textbf{0.52} \pm  \textbf{0.48} $ & $\textbf{5.77} \pm \textbf{3.00}$  \\    
  \bottomrule
  \label{table:sim2real}
  \end{tabular}}
  \vspace{-1em}
\end{table}

A quantitative error evaluation between simulation and reality is presented in \Cref{table:sim2real}. We show a significant improvement of residual physics over system identification, not only via a lower average error (decreased by \SI{60.3}{\%}) but also a consistently more robust performance through lower standard deviation in test trajectories.

We visualize one test trajectory in \Cref{fig:sim2real_beam} on the left. Our residual physics framework hereby helps to overcome the numerical damping problem and captures real-world dynamics better than SysID. We also plot the errors at each time step in \Cref{fig:sim2real_beam} on the right. Our framework reduces the error at each step and has a smaller deviation than the simulation based on SysID.

Further, we find that our model can accelerate simulations for stiff systems.  Our framework allows us to use suboptimal but non-stiff system parameters and correct the errors that these parameters induce \textit{via} the residual forces.  From \Cref{table:time}, we observe that achieving an accurate simulation using SysID slows down the simulation by approximately 3.12$\times$, yet for ResPhys, it is only 1.96$\times$, while achieving a much higher accuracy than SysID. We note that the hybrid simulation has additional overhead compared to the base simulation, due to the injected forces adding numerical stiffening.  

\begin{table}[h]
  \centering
  \vspace{0.5em}
  \setlength{\tabcolsep}{3.5 pt}
  \caption{Time benchmark of simulation with suboptimal parameters (original), optimal parameters (SysID), and suboptimal parameters with ResPhys. Statistics taken over 5 test trajectories.}
  \scalebox{1.}{
  \begin{tabular}{lc}
  \toprule
  Simulation & Time per Trajectory (\SI{}{s})\\ 
  \midrule
   Original Simulation & $0.711\pm0.002$\\
   System Identification & $2.218\pm0.005$ \\
   Residual Physics Simulation & $1.393\pm0.004$ \\
  \midrule
   Residual Network Inference & $0.112\pm0.002$ \\
  \bottomrule
  \label{table:time}
  \end{tabular}}
  \vspace{-2em}
\end{table}

\subsection{Sim-to-Real for Pneumatically-Actuated Soft Arm}

We further test our framework on SoPrA~\cite{Toshimitsu_2021}, which presents a more challenging scenario due to its increase in mesh size, pneumatic actuation, hard-to-model anisotropic fiber reinforcements around the actuation chambers, and a higher likelihood of hardware fabrication errors. We actuate the arm using random pressure sequences generated from a multivariate normal distribution. We tune the covariance of the distribution such that the resulting pressure trajectories are smooth.
Between the collection of each trajectory, we ensure that the system has returned to its unactuated steady state. We clip the commanded pressures to \SI{20}{\kilo\pascal} to avoid SoPrA inflating too much and prevent DiffPD from diverging without modeling fiber reinforcements. 

Similar to the sim-to-real beam experiments, we first perform rigid registration to transform motion marker data into our simulation environment and then optimize the augmented dataset \rev{by choosing $\lambda=10^{-5}$ in \Cref{eq:sim2real_op}}. However, different from the beam experiment, we now have non-zero pressure actuation forces $\mathbf{f}^{\mathrm{ext}}_t$ in the network input. These forces are computed from the pressure sequences and applied on the inner faces of the pneumatic chambers of the arm.

\begin{figure}[!b]
  \vspace{-1.5em}
  \begin{subfigure}[t]{0.485\columnwidth}
  \centering
    \includegraphics{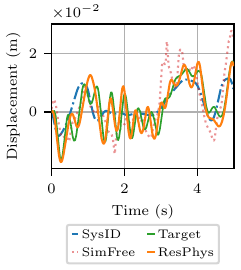}
  \end{subfigure}
  \hfill 
  \begin{subfigure}[t]{0.485\columnwidth}
  \centering
    \includegraphics[trim={2.5mm 5mm 0 0},clip]{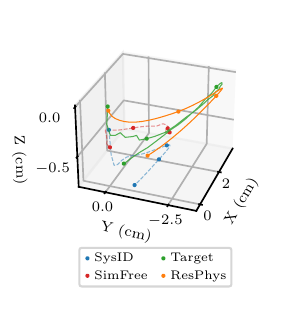}
  \end{subfigure}
  \caption{\rev{Sim-to-real results on the soft arm. (Left) Single 5s-excerpt test trajectory showing displacement in y-axis, averaged over motion markers. (Right) 3D displacement plot of the same trajectory between \SI{0}{s} and \SI{0.46}{s}, with points along the trajectory spaced \SI{0.15}{s} apart. For this particular segment of the trajectory, the average distance error for SysID is \SI{1.34}{mm}, for DD \SI{1.40}{mm}, and for ResPhys \SI{0.73}{mm}.}}
  \label{fig:sim2real_arm}
\end{figure}

\begin{figure}[!t]
  \begin{subfigure}[t]{0.485\columnwidth}
  \centering
    \includegraphics{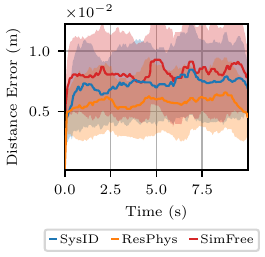}
  \end{subfigure}
  \hfill 
  \begin{subfigure}[t]{0.485\columnwidth}
  \centering
    \includegraphics{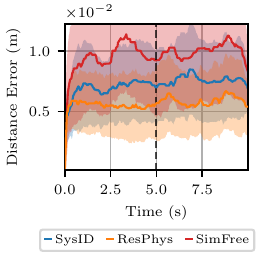}
  \end{subfigure}
  \caption{\rev{Mean and standard deviation over all test trajectories on the soft arm. For readability, curves are smoothened using a median filter with a window size of \SI{1}{s}. (Left) Train/test trajectories are both \SI{10}{s}. (Right) Train on \SI{5}{s} and test on extended \SI{10}{s}.}}
  \label{fig:arm_error}
\end{figure}

The final optimized SysID Young's modulus is \qty{237630}{\pascal} and Poisson's ratio \qty{0.4194}{}, but we observe little improvement in the error on test trajectories in \Cref{table:sim2real}. Using residual physics, the error is reduced by \SI{21.1}{\%}. The qualitative performance on an example trajectory is shown in \Cref{fig:sim2real_arm} and the average errors over all test trajectories in \Cref{fig:arm_error} on the left. \rev{We note the poor performance of the purely data-driven approach in this actuated robotic setup, highlighting the benefit of the underlying simulator for the residual physics approach in complex dynamical scenarios.}

Typically, machine learning models relying solely on data-driven approaches struggle to predict results in unseen domains accurately. We conduct an extrapolation experiment to test our framework's predictive capability on a longer time horizon. We keep all the training parameters and network architectures the same while only including each trajectory's first 500 time steps in the training set. Afterwards, we examine the model performance in predicting the last 500 time steps. Though prediction accuracy is decreased, our approach still outperforms both baselines as shown in \Cref{table:extrapolation} and \Cref{fig:arm_error} on the right. Our method shows stable long-time horizon predictions but with slowly increasing errors.

\begin{table}[h]
  \centering
  \setlength{\tabcolsep}{4.5 pt}
  \caption{\rev{Rollout error of markers for \SI{5}{s} (same length as training trajectories) and extrapolated to \SI{10}{s} (double the training length). Mean and standard deviation taken over all test trajectories.}}
  \scalebox{1.}{
  \begin{tabular}{lccc}
  \toprule
  Experiments & SysID (\SI{}{mm}) & SimFree (\SI{}{mm}) & ResPhys (\SI{}{mm})\\ 
  \midrule
  Test \SI{5}{s} & $7.51 \pm 3.63$ & $9.54 \pm 5.60$ & $5.79  \pm 3.09$ \\
  Test \SI{10}{s} & $7.57 \pm 3.59$ & $9.89 \pm 5.02$ & $6.10 \pm 3.22$ \\
  \bottomrule
  \label{table:extrapolation}
  \end{tabular}}
  \vspace{-2em}
\end{table}

\section{Conclusion and Future Work}

We demonstrated a hybrid residual physics framework for high-dimensional soft robots that combines numerical solvers with deep learning for residuals. By leveraging differentiable simulators and learned models, we eliminate the need for intricate domain-specific knowledge and full-state information about the robot. Instead, we shift our reliance to sparse marker data, simplifying the process for practitioners. Our framework helps reduce simulation errors under sim-to-sim and sim-to-real scenarios. We demonstrate its efficacy on passive and actuated soft structures such as a beam and pneumatic arm, showing that it consistently outperforms the system identification and data-driven baselines.

One drawback of our framework is the computationally expensive optimization procedure in the data pre-processing phase. This drawback, however, can be alleviated by more efficient simulators and parallel solving of independent problems. A second drawback is the generalization range of our method: if test data is sufficiently out of distribution, learned dynamics do not generalize.  In additional experiments, we found that generalization suffered on systems with 1.75$\times$ the internal actuation pressure. Future work should examine how to generalize beyond bounded training data and how to handle novel dynamical events, such as contact interactions.

Our results reveal some potential directions for future exploration. Since the network is provided a time sequence of motion data and we autoregressively call the network at inference time, we necessarily accumulate errors over longer trajectories. A data-driven model for long sequences would be suitable to address this problem, but due to the high DoFs of soft robots, it is hard to train the sequence model directly on the state of the soft robot. A powerful low-dimensional latent representation would provide a promising future avenue for investigation. Lastly, the question of generalizability should be addressed for varying shapes of soft robots. Until now, residual learning frameworks have been limited to single geometries. In future work, we hope to develop a single residual physics network applicable across various soft robot morphologies, proving the same level of applicability as numerical solvers while easing the need for laborious modeling.


\addtolength{\textheight}{-1.7cm}   


\bibliographystyle{IEEEtran}
\bibliography{bibliography}

\end{document}